\documentclass{l4dc2025}

\title[Asymptotics of Linear Regression with Linearly Dependent Data]{Asymptotics of Linear Regression with Linearly Dependent Data}
\usepackage{times}
\usepackage{wrapfig}

\coltauthor{\Name{Behrad Moniri} \Email{bemoniri@seas.upenn.edu}\\[0.1cm]
 \Name{Hamed Hassani} \Email{hassani@seas.upenn.edu}\\[0.1cm]
 \addr University of Pennsylvania, Philadelphia, PA, USA.}

\usepackage{hyperref}

\usepackage{amsmath,amsfonts,bm}

\def\eqref#1{equation~\ref{#1}}

\def\1{\bm{1}}

\DeclareMathAlphabet{\mathsfit}{\encodingdefault}{\sfdefault}{m}{sl}
\SetMathAlphabet{\mathsfit}{bold}{\encodingdefault}{\sfdefault}{bx}{n}

\newcommand{\E}{\mathbb{E}}

\newcommand{\R}{\mathbb{R}}

\newcommand{\bX}{\mathbf{X}}

\newcommand{\tr}{\operatorname{tr}}

\newcommand{\normal}{{\sf N}}
\newcommand{\ep}{\varepsilon}
\newcommand{\cD}{\mathcal{D}}

\newcommand*\colourcheck[1]{%
  \expandafter\newcommand\csname #1check\endcsname{\textcolor{#1}{\ding{52}}}%
}

\colourcheck{green}
\newcommand\independent{\protect\mathpalette{\protect\independenT}{\perp}}
\def\independenT#1#2{\mathrel{\rlap{$#1#2$}\mkern2mu{#1#2}}}

\newtheorem{condition}[theorem]{Condition}

\renewcommand{\epsilon}{\varepsilon}
\usepackage{cleveref}

\usepackage[margin=1in]{geometry}
\usepackage{ amsopn, bm, xcolor, amsmath}
\usepackage{multirow}
\usepackage{booktabs}
\usepackage{pifont}

\usepackage{microtype}
\usepackage{graphicx}
\usepackage{booktabs} 

\usepackage{amssymb}
\usepackage{mathtools}
\usepackage[utf8]{inputenc}

\def\hmath$#1${\texorpdfstring{{\rmfamily\textit{#1}}}{#1}}

\usepackage{setspace}

\begin{document}

\maketitle

\begin{abstract}%
In this paper we study the asymptotics of linear regression in settings with non-Gaussian covariates where the covariates exhibit a linear dependency structure, departing from the standard assumption of independence. We model the covariates using stochastic processes with spatio-temporal covariance and analyze the performance of ridge regression in the high-dimensional proportional regime, where the number of samples and feature dimensions grow proportionally. A Gaussian universality theorem is proven, demonstrating that the asymptotics are invariant under replacing the non-Gaussian covariates with Gaussian vectors preserving mean and covariance, for which tools from random matrix theory can be used to derive precise characterizations of the estimation error.  The estimation error is characterized by a fixed-point equation involving the spectral properties of the spatio-temporal covariance matrices, enabling efficient computation. We then study optimal regularization, overparameterization, and the double descent phenomenon in the context of dependent data. Simulations validate our theoretical predictions, shedding light on how dependencies influence estimation error and the choice of regularization parameters.
\end{abstract}


\section{Introduction}
Linear regression is one of the most widely used technique in machine learning and statistics. When given independent training samples, there are many tools available to study its performance under a set of relatively general conditions (see e.g., \cite{hsu2012random, oliveira2016lower, mourtada2022exact}, etc.). In particular, the asymptotic  behavior of the estimation error in this case has been fully characterized in the high-dimensional proportional regime where the number of samples and the covariate dimension tend to infinity with a proportional rate (see e.g., \cite{dobriban2018high,wu2020optimal,hastie2022surprises}, etc.). 

In contrast, much less is known when the independence assumption on training samples is lifted. The problem of linear regression with dependent data has a wide range of applications from system identification in linear dynamical systems to time-series analysis (see e.g., \cite{sysid,kailath2000linear,verhaegen2007filtering,chiuso2019system,tsiamis2023statistical}, etc.). Prior work has examined fairly general dependency structures, often providing finite-sample concentration inequalities on the estimation error of (ridgeless) linear regression without deriving precise formulas (see, e.g., \cite{simchowitz2018learning, faradonbeh2018finite, sarkar2019near, nagaraj2020least, ziemann2024noise}). 

Very recently, \cite{luo2024roti, atanasov2024risk} studied high-dimensional asymptotics of ridge regression with non-i.i.d. data under different assumptions for the covariates.  \cite{luo2024roti} studied the problem under the assumption that the covariates follow a right-rotationally invariant covariate distribution. \cite{atanasov2024risk} assumed that the covariates are generated according a \textit{Gaussian} stochastic process with \textit{spatio-temporal covariance}. The authors derived precise formulas for the asymptotic training and test errors and used it to derive generalized cross validation estimators (GCV) to predict the test error. 

In this paper, we also assume that the covariates are generated according a general stochastic process with a spatio-temporal covariance; lifting the Gaussianity assumption of prior work. In this limit, we prove that in this dependent data setting, a \textit{Gaussian Universality} property holds and the estimation error remains invariant under changing the covariates to Gaussian vectors while keeping their mean and covariance structure intact. Further, we provide novel precise formulas for the limiting estimation error. We then study the optimal ridge regularization, the effect of regularization and overparameterization, and also the double descent phenomenon in the setting with dependent data. In particular, we show that although the optimal ridge regularization is independent of the dependency structure, the dependence among covariates affects the degree of overparameterization in which the peak of the double descent and the phase transition from underparameterization to overparameterization happens.

\section{Preliminaries}
In this paper, we also consider the  study the problem of linear regression with \textit{linearly} dependent data. We assume that we observe $n$ samples of an input $d$-dimensional time-series $\{x_i\}_{i = 1}^{n}$, and an output real-valued time-series $\{y_i\}_{i = 1}^{n}$ from the following linear model
\begin{align}
    \label{eq:data_gen}
    y_i = \langle x_i, \beta_\star\rangle + \epsilon_i, \quad \text{for}\quad i = 1, 2, \dots, n,
\end{align}
where $\beta_\star \in \R^d$ is an unknown linear map, and $\{\ep_i\}_{i=1}^{n}$ is an additive noise independent of the input time-series $\{x_i\}_{i=1}^{n}$ with $\ep_i \overset{\mathrm{\rm i.i.d.}}{\sim} \normal(0, \sigma_\ep^2)$.  The goal is to use the samples $\{(x_i, y_i)\}_{i \in [n]}$ to recover the unknown vector $\beta_\star$.  To estimate $\beta_\star$, we will use the ridge estimator with parameter $\lambda \in \R$, given by  
\begin{align}
    \label{eq:ridge}
    \hat\beta &:= \arg\min_{\beta \in \R^d} \frac{1}{n} \sum_{i = 1}^{n} \left(y_i - \langle x_i, \beta \rangle\right)^2  + \lambda \Vert\beta\Vert_2^2.
\end{align}

We assume that $\{x_i\}_{i = 1}^{n}$ follows a stochastic process with \textit{Spatio-Temporal Covariance}, with a temporal covariance structure that is the same in all dimensions. Such dependency structures has been previously been studied in the literature. See e.g., \cite{nakakita2022benign,chen2021space,kyriakidis1999geostatistical}.
Formally, we let $X := \left[\,x_1\, |\, \cdots\, |\, x_n\, \right]^\top \in \R^{n \times d}$ and make the following assumption on  $X$.
\begin{condition}[Covariate Structure]
    \label{def:covariates}
    We assume that $X = A Z B$ in which $Z \in \R^{n \times d}$ is a random matrix with i.i.d. entries drawn from a distribution $\cD_z$, and that $A \in \R^{n\times n}$ and $B \in \R^{d \times d}$ are deterministic matrices controlling the temporal and spatial dependencies of the covariates.
\end{condition}
This dependency structure is fairly general. We will present some examples of problems satisfying this condition in Section~\ref{sec:examples}. Defining the output vector $y := [\,y_1, \dots, y_n\,]^\top \in \R^n$ and the noise vector $\ep = [\,\ep_1, \dots, \ep_n\,]^\top \in \R^n$, we can use \eqref{eq:data_gen} to write $y = X \beta_\star  + \ep$. With this matrix notation,  the estimator $\hat\beta$ is given by $\hat\beta = (X^\top X + \lambda n I_d )^{-1} X^\top y.$
We study this linear regression problem in the following asymptotic regime. The standard linear regression problem without dependencies has been studied extensively in the literature in a similar high-dimensional proportional regime (see e.g., \cite{tulino2004random,dobriban2018high,wu2020optimal,hastie2022surprises}, etc.). 
\begin{condition}[High-Dimensional Asymptotics]
\label{cond:high_dim}
    The sample size $n$ and the dimension $d$ tend to infinity at a proportional rate; i.e., $d, n \to \infty,  \text{ with }  d/n \to \gamma,$ where $\gamma>0$ is a constant. Throughout the paper, we will denote this limit by $\lim_{d/n \to \gamma}$.
\end{condition}
Consistent with a line of prior work (e.g., \cite{dobriban2018high,adlam2020understanding,adlam2020neural,disagreement}), we further make the following \textit{random-coefficient} assumption on the ground truth vector $\beta_\star$. We assume that the effect strength of each feature is drawn independently at random. Thus, our results can be viewed as an average-case analysis over parameters. 
\begin{condition}[Random Coefficient]
    \label{cond:random_effect}
     The coefficient $\beta_\star \in \R^d$ is random with $\beta_\star \sim \normal\left(0, \frac{\alpha^2}{d} I_d\right).$
\end{condition}
Note that based on this assumption, the vector the covariates $\{x_i\}$ do not carry any information about $\beta_\star$. Extending the results for the case where the regression coefficients are allowed to be anisotropic is an interesting problem, and we leave it as future work. See e.g., \cite{mel2021anisotropic,wu2020optimal,hastie2022surprises} for work in this direction in the i.i.d. setting.

To measure the performance of the ridge regularized linear regression estimator, in this paper we focus on the estimation error defined as
\begin{align}
    \label{eq:def_R}
    R(\hat\beta) := \|\hat\beta - \beta_\star\|_2^2.
\end{align}


\subsection{Related Work}
The study of linear regression with dependent data has attracted significant interest, particularly in the control theory community. The typical approach involves deriving non-asymptotic upper and lower bounds on the estimation error for linear regression under various dependency structures. For example, \cite{nagaraj2020least} examine this problem (without regularization) under a realizability condition, assuming data from an exponentially ergodic Markov chain. Additionally, \cite{simchowitz2018learning} showed that if we assume both realizability and that the noise forms a martingale difference sequence, then dependent linear regression is no harder than its independent counterpart. Meanwhile, \cite{ziemann2024noise} established upper bounds for random design linear regression with dependent ($\beta$-mixing) data, without imposing realizability assumptions. Another closely related problem, parameter identification in autoregressive models, has also been widely studied (see \cite{tsiamis2023statistical} for a recent survey).  

Studying problems with non-i.i.d. data is a recent trend in high-dimensional statistics. For example, \cite{bigot2024high} study the ridge regression in the high-dimensional regime with independent but non-identically distributed samples. \cite{zhang2024spectral} studies GLMs with correlated designs, and \cite{zhang2024matrix} study the problem of matrix denoising with correlated noise.

Most closely related to the current paper is the results of \cite{nakakita2022benign} and the very recent results of \cite{atanasov2024risk}. \cite{nakakita2022benign}  derive bounds for linear regression with a spatio-temporal covariance structure similar to the one we consider. Their analysis investigates the influence of the spectral properties of temporal covariance matrix on these bounds. \cite{atanasov2024risk} consider the problem of ridge regression trained on a dataset with a spatio-temporal dependency structure, but for Gaussian covariates. They analyze the precise asymptotics of in-sample and out-sample risks of the ridge estimator and propose a modified generalized cross validation to estimate the out-sample risk, given the in-sample risk. \cite{luo2024roti} solve a similar problem with the assumption that the design matrix $\bX$ is right-rotationally invariant. Although this setting does not restrict the covariates to be Gaussian, it cannot cover settings with general spatial covariance.

In this paper, we extend prior work by studying the estimation error for general \textit{non-Gaussian covariates} with spatio-temporal covariance by proving a Gaussian universality theorem. We derive a novel, precise asymptotic characterization of the estimation error, revealing how ridge regularization, the ratio $d/n$, and the spectral properties of the spatial and temporal covariance matrices jointly influence the estimation error. Additionally, we demonstrate novel phenomenon: while the optimal ridge regularization remains unaffected by the dependency structure of the covariates, the degree of overparameterization significantly impacts the location of the double descent peak and the phase transition from underparameterization to overparameterization.

\section{Main Results}
In this section, we state the main results of the paper that characterize the high-dimensional limits of the estimation error $R(\hat\beta) = \Vert\hat\beta - \beta_\star\Vert_2^2$.  Using the fact that $y = X \beta_\star + \ep$, the estimator $\hat\beta$ can be written as
\begin{align*}
    \hat\beta  = \beta_\star - \lambda n (X^\top X + \lambda n I_d)^{-1}\beta_\star + (X^\top X + \lambda n I_d)^{-1}X^\top \ep.
\end{align*}
Thus, plugging this into \eqref{eq:def_R}, the estimation error $R(\hat\beta)$ satisfies
\begin{align*}
    R(\hat\beta) = \underbrace{\ep^\top X(X^\top X + \lambda n I_d)^{-2}X^\top \ep}_{:= \mathcal{V}} + \underbrace{\lambda^2 n^2 \beta_\star^\top (X^\top X + \lambda n I_d)^{-2} \beta_\star}_{:=\mathcal{B}} - 2\lambda n\beta_\star^\top (X^\top X + \lambda n I_d)^{-2}X^\top \ep.
\end{align*}
This is the bias-variance decomposition for $R(\hat\beta)$ where $\mathcal{V}$ is the variance and $\mathcal{B}$ is the bias. Using the Hanson-Wright inequality \citep{rudelson2013hanson}, and the fact that $\ep$ is independent of all other randomness in the problem, the third term in the above sum converges to zero  and we have $|R(\hat \beta) - \bar{\mathcal{B}} - \bar{\mathcal{V}}| \to 0$ in which the values $\bar{\mathcal{V}} := \E_\ep[\mathcal{V}]$ and $\bar{\mathcal{B}} := \E_{\beta_\star}[\mathcal{B}]$ are given by
\begin{align*}
     \bar{\mathcal{V}} := \frac{\sigma_\ep}{n} \; \mathrm{trace}\left[X(X^\top X + \lambda n I_d)^{-2}X^\top\right], \quad \text{and}\quad \bar{\mathcal{B}} = \frac{\alpha^2\lambda^2 n^2}{d} \; \mathrm{trace} \left[(X^\top X + \lambda n I_d)^{-2} \right].
\end{align*}
Next, we define the functional $m: \R \times \R \to \R$, as
\begin{align}
    \label{eq:m_def}
    m(\lambda; \gamma) :=  \, \lim_{d/n \to \gamma} \frac{1}{d}\tr\left( (B^\top Z^\top A^\top AZB + \lambda I_d )^{-1}\right).
\end{align}
Recalling that under Condition~\ref{cond:high_dim}, $X  = AZB$, we have 
    \begin{align}
    \label{eq:in_terms_of_m}
        \left|\bar{\mathcal{B}} + \alpha^2 \lambda^2 \frac{\partial m(\lambda;\gamma)}{\partial \lambda} \,\right| \to 0, \quad\hspace{-0.2cm} \text{and}\quad \hspace{-0.2cm}\left|\bar{\mathcal{V}} - \gamma  \sigma_\ep^2 m(\lambda; \gamma) - \gamma \sigma_\ep^2 \lambda \frac{\partial m(\lambda;\gamma)}{\partial \lambda} \,\right| \to 0.
\end{align}    
Thus, the estimation error $R(\hat\beta)$, variance $\bar{\mathcal{V}}$, and bias $\bar{\mathcal{B}}$ are fully characterized by the function $m(\lambda; \gamma)$. For this reason, we will focus on studying the function $m(\lambda; \gamma)$ in the subsequent sections.


\subsection{Universality}
\label{sec:universality}
Under Condition~\ref{def:covariates}, the covariates are given by \( X = A Z B \) in which \( Z \in \mathbb{R}^{n \times d} \) is a random matrix with i.i.d. entries drawn from a distribution \( \mathcal{D}_z \), and \( A \in \mathbb{R}^{n \times n} \) and \( B \in \mathbb{R}^{d \times d} \) are deterministic matrices. In this section, we show that for a broad class of distributions \( \mathcal{D}_z \), under the high-dimensional setting of Condition~\ref{cond:high_dim}, the limiting value of the estimation error \( R(\hat{\beta}) \) is invariant under replacing the distribution \( \mathcal{D}_z \) with a Gaussian distribution with the same mean and variance as \( \mathcal{D}_z \). This statement is formalized in the following theorem.

\begin{theorem}[Gaussian Universality]  
\label{thm:universality}
Let $G= [g_1, \dots, g_n]^\top \in \R^{n \times d}$ have entries drawn i.i.d. from  Gaussian distribution with the same mean and variance as $\cD_z$, and let $\tilde X = A G B$ and  $\tilde y = \tilde X \beta_\star + \ep$. Under Conditions~\ref{def:covariates}-\ref{cond:random_effect}, for the ridge estimator given the dataset $\tilde X, \tilde y$, i.e., $\tilde \beta = (\tilde X^\top \tilde X + \lambda n I_d )^{-1} \tilde X^\top \tilde y$, we have
\begin{align*}
    |R(\hat \beta) - R(\tilde \beta)| \to 0,
\end{align*}
with probability $1 - o(1)$, under the condition that $\cD_z$ is sub-Gaussian.
\end{theorem}
This theorem shows that under mild conditions, the limiting value of the estimation error depends on the first two moments of $\cD_z$ and not on the other properties of the distribution. 
\paragraph{Proof Sketch.} To prove this theorem, given \eqref{eq:in_terms_of_m}, it is enough to show that $m(\lambda;\gamma)$ is invariant under the change of $\cD_z$ to the corresponding Gaussian. For this, we use a Lindeberg exchange argument \citep{lindeberg1922neue,korada2011applications,abbasi2019universality} where we replace the rows of $Z$ with corresponding Gaussian vectors one at a time and show that the error incurred by these replacements are negligible.  Let $G= [g_1, \dots, g_n]^\top \in \R^{n \times d}$ be a matrix with entries drawn i.i.d. from  Gaussian distribution with the same mean and variance as $\cD_z$. For any $i \in [n]$  we define the matrix $M_i$ as
\begin{align*}
    M_i := \sum_{k = 1}^{i - 1} e_k g_k^\top + \sum_{k = i+1}^{n} e_k z_k^\top,
\end{align*}
where $e_1, \dots, e_n$ are the standard basis of $\R^n$. The matrix will be used to interpolate between $Z$ and $G$.  Also, we define $Z_i := M_i + e_i g_i^\top$. Note that $G = Z_n$ and $Z = Z_0$.  Denoting $\Omega_A = A^\top A$, we can write  
\begin{align*}
    \Delta := \Big|\mathrm{trace}\left(B^\top G^\top \Omega_A G B + \lambda n I_d\right)^{-1} &- \mathrm{trace}\left(B^\top Z^\top \Omega_A Z B + \lambda n I_d\right)^{-1}\Big| \nonumber \\
    &\hspace{-3cm}\leq \sum_{i = 1}^{n}\Big|\,\mathrm{trace}\left(B^\top Z_i^\top \Omega_A Z_i B + \lambda n I_d\right)^{-1} - \mathrm{trace}\left(B^\top Z_{i-1}^\top \Omega_A Z_{i-1} B + \lambda n I_d\right)^{-1} \Big|.
\end{align*}
To complete the proof of the theorem, it is enough to show that each term in the above sum is $o(1/n)$. For this, we use that fact that $Z_i = M_i + e_i g_i^\top$ and  $Z_{i - 1} = M_i + e_i z_i^\top$ and the apply the Woodbury matrix identity (alongside some elementary algebra) to arrive at 
\begin{align*}
    \Delta \leq \sum_{i = 1}^{n}\Big|f_i(g_i) - f_i(z_i) \Big|,
\end{align*}
in which for any $u \in \R^d$, the function $f_i(u)$ is given by
\begin{align*}
    f_i(u) := \mathrm{trace}\left(\begin{bmatrix}
         u^\top B\, S_i\ B^\top u & 1 + u^\top B\, S_i\,B^\top  v_i\\[0.2cm]
        1 +  v_i^\top B\, S_i \,B^\top u & { v_i^\top B\, S_i \, B^\top v_i- e_i^\top \Omega_A e_i}
    \end{bmatrix}^{-1} \begin{bmatrix}
        u^\top B\,  S_i^2\,B^\top u  & u^\top B\, S_i^2\, B^\top v_i\\[0.3cm]
        u^\top B \, S_i^2\,B^\top v_i & v_i^\top B\, S_i^2 \, B^\top v_i
    \end{bmatrix}\right),
\end{align*} 
where $S_i := \left(B^\top M_i^\top \;\Omega_A\; M_i B + \lambda n I_d\right)^{-1}$ and $v_i :=  M_i^\top \Omega_A e_i \in \R^d$.  Given that $\cD_z$ is sub-Gaussian and using the Hanson-Wright inequality and noting that by construction $ g_i, z_i \independent S_i$, for both $u = g_i$ and $u = z_i$ we have 
$u^\top B\,S_iB^\top \, u = \tr(B\,S_iB^\top) + O(1/\sqrt{n})$, and $v_i^\top B\,S_i B^\top  u = O(1/\sqrt{n})$. Similarly, $u^\top B\,S_i^2B^\top \, u = \tr(B\,S_i^2B^\top) + O(n^{-3/2})$, and $v_i^\top B\,S_i^2 B^\top  u = O(n^{-3/2})$. Also, using a simple order-wise argument $v_i^\top S_i \, v_i = O(1)$.  This shows that given $\omega_i = O(1)$, we get
\begin{align*}
    \left|f_i(g_i) - f_i(z_i)\right| = O(n^{-3/2}) = o(1/n),
\end{align*}
proving the theorem. The detailed proof can be found in Section~\ref{sec:pf:universality}. \hspace{6cm} $\blacksquare$

\subsection{Precise Asymptotics}
\label{sec:precise}
In this section, we provide a precise formula for $\lim_{d/n \to \gamma} m(\lambda;\gamma)$ and use it to fully characterize the estimation error $R(\hat\beta)$ in the high-dimensional setting of Condition~\ref{cond:high_dim}. For this, we assume that the empirical spectral distributions of $A^\top A$ and $B^\top B$ converge to limiting distributions $\mu_A$ and $\mu_B$ respectively and then write $m(\lambda;\gamma)$ in terms of the spectral properties of these limiting distributions.
\begin{condition} [Empirical Spectral Distribution] 
\label{cond:limiting_AB}
Let $\lambda_1, \dots, \lambda_n$ be the eigenvalues of $A^\top A \in \R^{n\times n}$ and $\tilde \lambda_1, \dots, \tilde \lambda_d$ be the eigenvalues of $B^\top B \in \R^{d\times d}$. We assume that as $n, d \to \infty$, we have
\begin{align*}
    \frac{1}{n} \sum_{i = 1}^{n} \delta_{\lambda_i} \implies \mu_A, \quad \text{and}\quad \frac{1}{d} \sum_{i = 1}^{d} \delta_{\tilde\lambda_i} \implies \mu_B,
\end{align*}
where $\mu_A$ and $\mu_B$ are two probability measures over $\R$, and $\implies$ denotes weak convergence.
\end{condition}
This is a mild technical condition and is standard in high-dimensional statistics. Under this condition, in the next theorem we provide a precise formulae for $\lim_{d/n \to \gamma} m(\lambda; \gamma)$, which we will later use to characterize the limiting value of the estimation error $R(\hat \beta)$.

\begin{theorem}
\label{thm:error_with_fixed_point}
Assume that the conditions of Theorem~\ref{thm:universality} hold. The function $m(\lambda;\gamma)$ converges in probability to $\bar m(\lambda;\gamma)$ that is given by $\bar m(\lambda;\gamma) = \kappa \, m_B(\kappa)/ \lambda$ where $\kappa$ is the solution of the following nonlinear equation,
\begin{align}
\label{eq:fixed_points}
    \lambda m_A \left(\frac{\lambda}{\gamma \kappa - \gamma \kappa^2 \, m_B(\kappa)}\right) + \gamma \kappa \left(\kappa\, m_B(\kappa)-1\right) \left(1 - \gamma + \gamma \kappa\, m_B(\kappa)\right) = 0,
\end{align}
where for any $z \in \R$, $m_A(z)$ and $m_B(z)$ are defined as
\begin{align*}
    m_A(z) := \int_\R \frac{1}{z+x} d\mu_A(x), \quad \text{and}\quad  m_B(z) := \int_\R \frac{1}{z+x} d\mu_B(x).
\end{align*}
\end{theorem}
This theorem shows that although the limiting expression for $\bar m(\lambda;\gamma)$ does not have a closed form, it can be written in terms of a single scalar $\kappa$ that is the solution to a scalar fixed-point equation which can be solved very efficiently using fixed-point iteration. Whereas directly using $R(\hat\beta) = \|\hat\beta - \beta_\star\|_2^2$ requires inversion and multiplication of large-dimensional matrices.

\paragraph{Proof Sketch.}  Given Theorem~\ref{thm:universality}, without loss of generality we assume that $\cD_z$ is Gaussian. To find the limiting value, we construct linear pencils  and use the theory of operator-valued free probability to derive the limit of $m(\lambda;\gamma)$. This technique has be  used to study random feature models (see e.g., \cite{adlam2020understanding,adlam2019random,tripuraneni2021covariate,mel2021anisotropic, disagreement, dohmatob2024strong}, etc.).  See \cite[Part 1]{bodin2024random} for an overview of the linear pencil technique in random matrix theory. 

A linear pencil is a block-matrix whose blocks are matrices with known spectral properties, and one of the blocks of its inverse is the desired function $m(\lambda;\gamma)$. Let $H_0 :=  \left({B^\top Z^\top {\Omega_A}\, Z\, B}/{\lambda n} +  I_d\right)^{-1}$, where $m(\lambda;\gamma) = \frac{1}{\lambda d}\tr(H_0).$ We can linearize the problem by introducing auxiliary random matrices $H_1, H_2, H_3, H_4$ to write
{\small
\begin{align}
    \underbrace{\begin{bmatrix}
        I & {0} & {0} & {0} & B^\top\\[0.2cm]
        -B & I & {0} & {0} & {0}\\[0.2cm]
        {0} & -Z/\sqrt{\lambda n} & I & {0} & {0} \\[0.2cm]
        {0} & {0} & -{\Omega_A} & I & {0} \\[0.2cm]
        {0} & {0} & {0} & -Z^\top/\sqrt{\lambda n} & I
    \end{bmatrix}}_{:= {L}}
    \begin{bmatrix}
        H_0\\[0.2cm]
        H_1\\[0.2cm]
        H_2\\[0.2cm]
        H_3\\[0.2cm]
        H_4
    \end{bmatrix} 
    = 
    \begin{bmatrix}
        I\\[0.2cm]
        {0} \\[0.2cm]
        {0} \\[0.2cm]
        {0} \\[0.2cm]
        {0} 
    \end{bmatrix}.
\end{align}
}
This shows that the matrix $H_0$ is equal to the top left block of ${L}^{-1}$. 
Hence, the matrix ${L}$ is a suitable linear pencil for our problem. The matrix $L$ can be written as $L = I - L_Z - L_{A, B},$ in which
{\small\begin{align*}
    L_Z := \begin{bmatrix}
        {0} & {0} & {0} & {0} & {0}\\[0.2cm]
        {0} & {0} & {0} & {0} & {0}\\[0.2cm]
        {0} & Z/\sqrt{\lambda n} & {0} & {0} & {0} \\[0.2cm]
        {0} & {0} & {0} & {0} & {0} \\[0.2cm]
        {0} & {0} & {0} & Z^\top/\sqrt{\lambda n} & {0}
    \end{bmatrix}, \quad \text{and} \quad L_{A, B} := \begin{bmatrix}
        0 & {0} & {0} & {0} & -B^\top\\[0.2cm]
        B & 0 & {0} & {0} & {0}\\[0.2cm]
        {0} & {0} &0 & {0} & {0} \\[0.2cm]
        {0} & {0} & {\Omega_A} & 0 & {0} \\[0.2cm]
        {0} & {0} & {0} & {0} & 0
    \end{bmatrix}.
\end{align*}}
We next define the matrix $\mathcal{G} := (\mathrm{id}\otimes\E \bar\tr) ( L^{-1}) = (\mathrm{id}\otimes\E \bar\tr) \left((I - L_Z -  L_{A, B})^{-1}\right) \in \R^{5 \times 5}$ where $(\mathrm{id}\otimes\E \bar\tr)$ is the normalized block-trace; i.e., replacing each block by its trace divided by the size of the block. This matrix can be seen as the operator-valued Cauchy transform \cite[Definition 9.10]{mingo2017free} of $ L_Z +  L_{A, B}$ evaluated at $I$; i.e., $\mathcal{G}_{ L_Z +  L_{A, B}} (I).$  The Cauchy transform of sum of random matrices has been studied extensively in the literature. A common tool for compute this is the R-transform introduced by  \cite{voiculescu1986addition,voiculescu2006symmetries} which enables the characterization of the spectrum of a sum of asymptotically freely independent random matrices. Because $ L_{A, B}$ is a deterministic matrix, the matrices $ L_Z$ and $ L_{A, B}$ are asymptotically freely independent \cite[Theorem 4.8]{mingo2017free} which enables us to use  the subordination property \cite[Equation 9.21]{mingo2017free} and write $\mathcal{G} = \mathcal{G}_{ L_Z +  L_{A, B}} (I)$ as
\begin{align}
 \label{eq:subord}
    \mathcal{G} = (\mathrm{id}\otimes\E \bar\tr) \left((I  - \mathcal{R}_{ L_{Z}}(\mathcal{G})  -  L_{A, B})^{-1}\right),
\end{align}
where $\mathcal{R}_{L_{Z}}(\mathcal{G})$ is the operator-valued R-transform \cite[Definition 9.10]{mingo2017free} of the matrix $L_{Z}$, evaluated at $\mathcal{G}$. If we can write the R-transform $\mathcal{R}_{L_{Z}}(\mathcal{G})$ as a function of $\mathcal{G}$, \eqref{eq:subord} will give a fixed-point equation for $\mathcal{G}$. Since the matrix $L_Z$ consists of i.i.d. Gaussian blocks, we can use \cite[Theorem 4]{far2006spectra} to compute $\mathcal{R}_{ L_{Z}}(\mathcal{G})$ in terms of $\mathcal{G}$. With this, we find that $\mathcal{R}_{ L_{Z}}( \mathcal{G})$ is a block matrix and its non-zero blocks are $[{\mathcal{R}_{L_Z}(\mathcal{G})}]_{3,4} = \frac{\gamma}{\lambda} [\mathcal{G}\,]_{2,5}, \quad \text{and}\quad [{\mathcal{R}_{L_Z}(\mathcal{G})}\,]_{5,2} = \frac{1}{\lambda} [\mathcal{G}]_{4,3}.$
Note that by construction, we have $ [\mathcal{G}\,]_{1,1} = \frac{1}{d}\tr(H_0) = \lambda m(\lambda;\gamma)$. Plugging this R-transform into \eqref{eq:subord}, and looking at the $(1,1)$, $(4,3)$, and $(2, 5)$ blocks of both sides, we get the following three equalities
\begin{align*}
        &\lambda m(\lambda;\gamma) = [\mathcal{G}\,]_{1,1} =  \frac{\lambda}{d}\tr \left((\lambda I + [\mathcal{G}\,]_{4,3} \,\Omega_B)^{-1}\right), \quad 
        [\mathcal{G}\,]_{4,3} = \frac{\lambda}{n}  \,\tr \left(A^\top A(\lambda I - \gamma\, \,[\mathcal{G}\,]_{2,5}\, \Omega_A)^{-1}\right),\quad\text{and}\\[0.1cm]
        &[\mathcal{G}\,]_{2,5} = -\frac{\lambda}{d}\, \tr\left(B^\top B (\lambda I + [\mathcal{G}\,]_{4,3}\,\Omega_B)^{-1}\right).
\end{align*}
Defining $\kappa := \lambda/[\mathcal{G}]_{4,3}$ and  simplifying this system of three equations, we arrive $m(\lambda; \gamma) = \kappa m_B(\kappa)/\lambda,$ and \eqref{eq:fixed_points}
 which concludes the proof. The full proof can be found in Section~\ref{sec:pf:error_with_fixed_point}. \hspace{4cm} $\blacksquare$

This theorem can be used alongside \eqref{eq:trace_expression} to find a formula for the limiting bias, variance, and estimation error in terms of the implicit variable $\kappa$ as follows.
\begin{corollary}
\label{cor:final}
Under the conditions of Theorem~\ref{thm:error_with_fixed_point}, with probability $1 - o(1)$, we have   
\begin{align}
    \label{eq:trace_expression}
    \bar{\mathcal{B}} \to - \alpha^2 \lambda^2 \frac{\partial \bar m}{\partial \lambda},\quad \bar{\mathcal{V}} \to \gamma  \sigma_\ep^2 \kappa \, m_B(\kappa)/ \lambda + \gamma \sigma_\ep^2 \lambda \frac{\partial \bar m}{\partial \lambda}, \quad \text{and} \quad |R(\hat\beta) - \bar{\mathcal{B}} - \bar{\mathcal{V}}| \to 0,
\end{align}   
in which $\kappa$ and $\bar{m}$ is defined in Theorem~\ref{thm:error_with_fixed_point} and $\frac{\partial \bar m}{\partial \lambda}$ can be derived by implicit differentiation of \eqref{eq:fixed_points}. The formula for $\frac{\partial \bar m}{\partial \lambda}$ in terms of $\kappa$ can be found in Section~\ref{sec:derivative}.
\end{corollary}
\subsection{Optimal Regularization}
\label{sec:optimal}
Despite the lack of a closed form solution for the limiting estimation error, the optimal value for the ridge parameter $\lambda$ can still be derived, even in the case where the covariates are dependent.
\begin{theorem}
    \label{thm:optimal}
    Let the data be generated according to \eqref{eq:data_gen}  and $\hat\beta$ be the solution to \eqref{eq:ridge} with regularization parameter $\lambda\in \R$. Assume that Condition~\ref{cond:random_effect} holds. The optimal value of the ridge regularization parameter is given by
    \begin{align*}
        \lambda_\star := \arg\min_{\lambda \in \R} \left(\lim_{d/n \to \gamma} R(\hat\beta)\right) = \sigma_\ep^2\gamma/\alpha^2.
    \end{align*}
\end{theorem}
The case where the covariates are i.i.d. was studied by \cite{dobriban2018high} where it was shown that the optimal value of $\lambda$ is again equal to $\sigma_\ep^2\gamma/\alpha^2$. Hence, the optimal  $\lambda$ does \textit{not} depend on the dependency structure of the covariates.
\paragraph{Proof.} Similar to \cite{dobriban2018high}, we note that in the finite $d, n$ setting and under the assumption that $\ep \sim \normal(0, \sigma_\ep^2 I_n)$ and $\beta_\star\sim \normal(0, \alpha^2 I_d/d)$, the ridge regression of \eqref{eq:ridge} with $\lambda = \lambda_n \sigma_\ep^2d/\alpha^2n$ is the Bayes-optimal estimator for $\beta_\star$. Using \cite[Lemma 6.1]{dobriban2018high}, $\lambda_\star = \lim_{d/n \to \gamma} \lambda_n = \sigma_\ep^2\gamma/\alpha^2$ is the optimal ridge regularization under the high-dimensional limit of Condition~\ref{cond:high_dim}.

\section{Special Cases}
In this section, we study special cases of Corollary~\ref{cor:final} where the fixed point equation of \eqref{eq:fixed_points} simplifies. Although \eqref{eq:fixed_points}  is a complicated fixed point equation, in special cases it can be used to recover prior work that study special cases of this problem.

\paragraph{Case 1: $B^\top B = I_d$ and $A^\top A = I_n$}  In this case, for any $z \in \R$ we have $m_{B}(z) = m_{A}(z) = (1+z)^{-1}$. Using the fixed-point \eqref{eq:fixed_points}, we find that $\bar m(\lambda;\gamma)$ is the solution of $\lambda \gamma \bar{m}^2(\lambda; \gamma) +  \left(1 + \lambda - \gamma\right) \bar{m}(\lambda; \gamma)  - 1 = 0$, which corresponds to the standard formula for the Stieltjes transform of the Marchenko-Pastur law \citep{bai2010spectral}. This recovers the formula for $\bar{m}$ in the case where samples are i.i.d. and have a covariance matrix equal to identity \citep{tulino2004random}.

\paragraph{Case 2:  $A^\top A = I_n$} In this case, for any $z \in \R$ we have $m_{A}(z) = (1+z)^{-1}$. Plugging this into the fixed-point equations in \eqref{eq:fixed_points}, we find that $\bar m(\lambda; \gamma) = \kappa m_{B}(\kappa)/\lambda$ where $\kappa$ satisfies $\gamma\, \kappa^2 \,m_{B}(\kappa) + \kappa \left(1 - \gamma\right) - \lambda = 0.$
In this case, as expected, we recover the formula for the i.i.d. case (but with a general covariance matrix) from  \cite{dobriban2018high,hastie2022surprises,wu2020optimal}.

\paragraph{Case 3: $B^\top B = I_d$} In this case, for any $z \in \R$ we have $m_{B}(z) = (1+z)^{-1}$. Plugging this into the fixed-point equations in \eqref{eq:fixed_points}, we find that $\bar m(\lambda; \gamma)$ is the solution to the following equation
\begin{align*}
       \lambda \gamma^2\, \bar m^2(\lambda; \gamma) + \gamma\, (1 - \gamma) \,\bar m(\lambda; \gamma) -m_{A}\left(\frac{1}{\gamma \,\bar m(\lambda; \gamma)}\right) = 0.
\end{align*}

\subsection{Examples}
\label{sec:examples}
The linear dependency structure we assume for the covariates in Condition~\ref{def:covariates} is fairly general. To illustrate this, we present two examples of linear regression problems that fit within the framework of this paper. First, given $\omega, \omega_0, \dots, \omega_q \in \R$ we define the following two Toeplitz matrices (all other entries are zero)
{\small
\begin{align}
\label{eq:examples}
    A_{\rm AR} =   
\begin{bmatrix}
    \omega_0    &        &        &        &         \\
    \vdots & \omega_0    &        &        &         \\
    \omega_q    & \vdots & \ddots &        &         \\
           &  \omega_q   & \ddots &    \omega_0 &         \\
           &        &  \omega_q   & \dots  &    \omega_0  \\
\end{bmatrix} \in \R^{n \times n},\quad \text{and} \quad 
    A_{\rm R} =   
\begin{bmatrix}
    1  &            &        &        &         \\
    0  & \omega     &        &        &         \\
       & 1 - \omega &    1   &        &         \\
       &            &    0   &    \omega  &         \\
       &            &        & 1 - \omega  &    \ddots  \\
       &            &        &             &    \ddots  \\
\end{bmatrix}\in \R^{n \times n}.
\end{align}}

\noindent\textbf{Example 1 (Autoregressive):}  Let $z_1, \dots, z_n \overset{\rm i.i.d.}{\sim} \normal(0, I_d)$, and $\{x_i\}_{i = 1}^{n}$ and $\{y_i\}_{i = 1}^n$ be generated according from the following autoregressive  model with order $q \in \mathbb{N}$, given by
\begin{align*}
    x_t = \sum_{i = 0}^{q} \omega_{i} z_{t-i}, \quad \text{and} \quad y_t = \langle x_t, \beta_\star \rangle + \ep_i,
\end{align*}
 where $\omega_0, \dots, \omega_q \in \R$ are real coefficients. This data generation process satisfies Condition~\ref{def:covariates} with $B = I_d$, and $A$ equal to the Toeplitz matrix $A_{\rm AR}$ from \eqref{eq:examples}.

The limiting spectral distribution of such Toeplitz converges and the limiting distribution is characterized by the celebrated Szegő theorem (see e.g, \cite{GrenanderSzego1958, tyrtyshnikov1996unifying,gray1972asymptotic}). In particular, it is known this example satisfies Condition~\ref{cond:limiting_AB} and for any $z \in \R$, we have
\begin{align*}
    m_{A_{\rm AR}}(z) = \frac{1}{2\pi}\int_{0}^{2\pi} \frac{1}{z + |f|^2(\lambda)}\; d\lambda, \quad \text{in which}\quad f(\lambda) := \sum_{k = 0}^q \, \omega_k e^{ik\lambda}.
\end{align*}
Hence, Corollary~\ref{cor:final} fully characterizes the bias, variance, and estimation error for this example.

\noindent\textbf{Example 2 (Redundancy):} In this example, we will present a dependency structure that exhibit a controllable amount of redundancy, and we will use this setting for some experiments in the next section. Let $\omega \in [0,1]$ be a real number and  $z_1, \dots, z_{n} \overset{\rm i.i.d.}{\sim} \normal(0, I_d)$. Assume that the covariates are given by $x_i = z_i$ if $i$ is odd and $x_i = \omega z_{i} + (1-\omega)z_{i-1}$ if $i$ is even and that $y_i$ are generated similar to Example 1. The case where $\omega = 1$ corresponds to the case with i.i.d. $\normal(0,I_d)$ covariates, and the case with $\omega = 0$ corresponds to the case where each covariate is repeated twice. This data generation process satisfies Condition~\ref{def:covariates} with $B = I_d$, and $A$ equal to $A_{\rm R}$ from \eqref{eq:examples}.

\begin{figure}[t]
    \centering
    \includegraphics[width=0.95\linewidth]{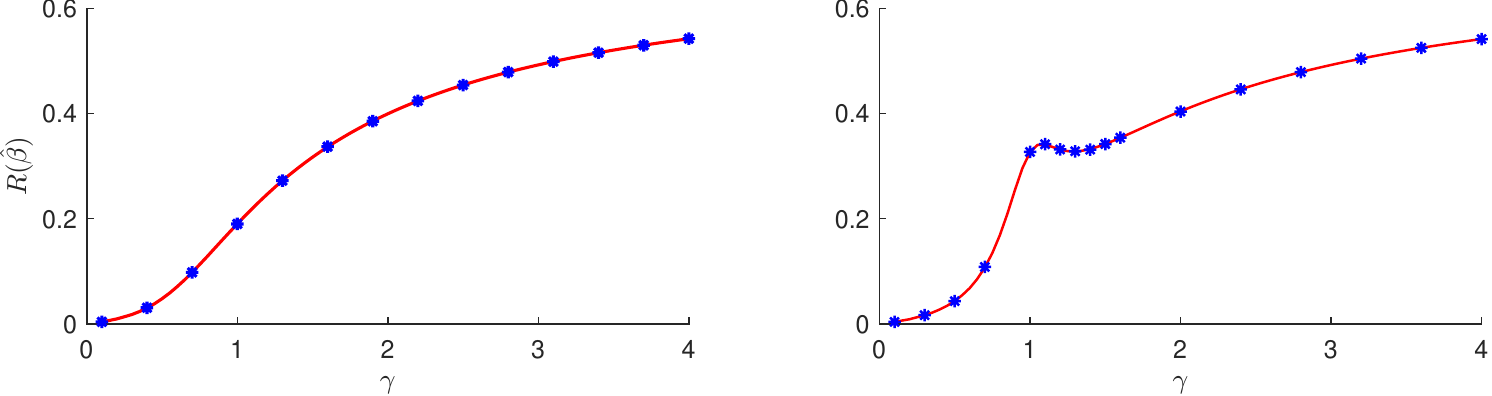}
    \caption{The estimation error $R(\hat\beta)$ as a function of $\gamma$ with $\mu_A = 1/3\, \delta_{1} + 1/3\, \delta_{2} + 1/3\, \delta_{3}$, $\mu_B = 1/2\, \delta_{1} + 1/2\, \delta_{2}$, $\sigma_\ep = 0.2$, and $\alpha = 0.7$.  We set $\lambda = 0.03$ (Left) and $\lambda = \gamma$ (Right). In these plots, the solid lines are the theoretical predictions using Corollary~\ref{cor:final} and the dots are numerical predictions, averaged over 50 trials. }
    \label{fig:theory_vs_experiments}
\end{figure}
\section{Experiments and Simulations}
\label{sec:sim}

\begin{figure}
    \centering
    \includegraphics[width=0.95\linewidth]{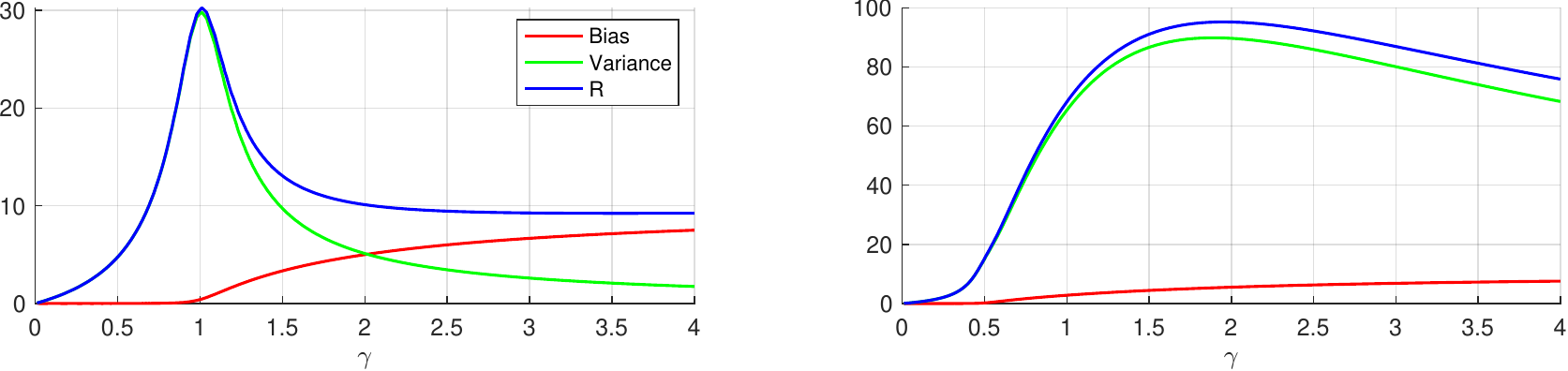}
    \caption{The estimation error, bias, and variance of Example 2 in Section~\ref{sec:examples} with $\sigma_\ep = 1, \alpha = 1,$ and $\gamma = 0.2$ as a function of $\gamma$ for $\omega = 0.8$ (Left) and $\omega = 0.2$ (Right).}
    \label{fig:double}
\end{figure}

\paragraph{Theory vs. Simulation.} We set $\mu_A = 1/3\, \delta_{1} + 1/3\, \delta_{2} + 1/3\, \delta_{3}$, $\mu_B = 1/2\, \delta_{1} + 1/2\, \delta_{2}$, $\sigma_\ep = 0.2$, and $\alpha = 0.7$. In this experiment, we consider the estimation error once with a fixed $\lambda = 0.03$, and once with $\lambda = \gamma$. In these plots, we use Corollary~\ref{cor:final} to plot the limiting $R(\hat\beta)$ as a function of $\gamma \in [0,4]$. For each setting, we also numerically simulate results using \eqref{eq:def_R} and average over 50 trials.  The results can be found in Figure~\ref{fig:theory_vs_experiments}.   We observe that the theoretical predictions of the paper match perfectly with the numerical simulations. 
\paragraph{Effect of Dependence.} We consider the setting of Example 2 in Section~\ref{sec:examples} and set $\sigma_\ep =1, \alpha = 1$, and $\lambda = 0.05$. In Figure~\ref{fig:double}, we plot $R(\hat\beta)$, variance, and bias, once with $\omega = 0.2$ (Right) and once for $\omega = 0.8$ (Left).

\newpage

\begin{wrapfigure}{r}{0.43\textwidth}
  \begin{center}
    \includegraphics[width=0.85\linewidth]{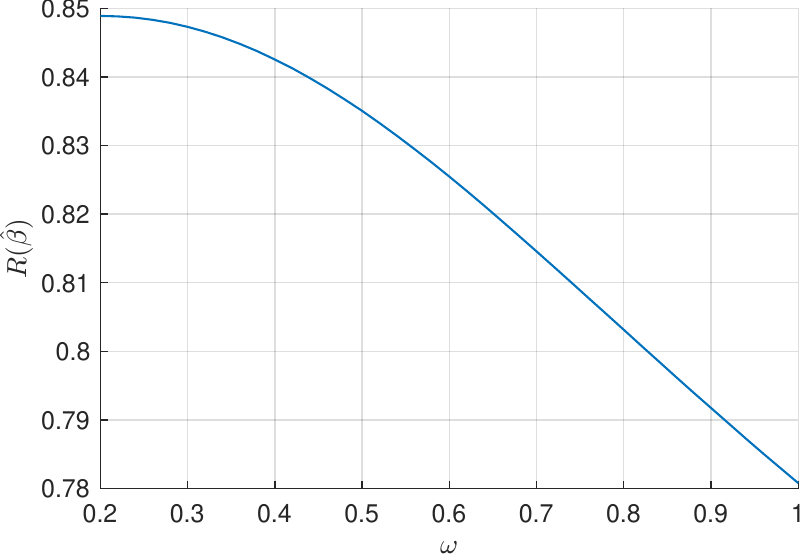}
    \caption{The estimation error as a function of $\omega$ for Example 2 in Section~\ref{sec:examples} as a function of $\omega$ with  $\sigma_\ep =1, \alpha = 1$, $\gamma = 2$ and $\lambda = \lambda_\star$.\vspace{0.2cm}}
    \label{fig:as_dep}
    \includegraphics[width=0.85\linewidth]{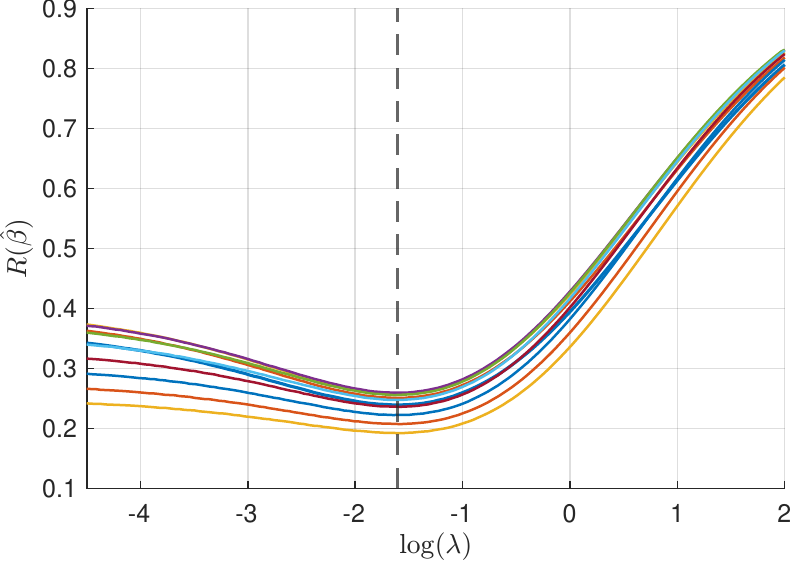}
    \caption{The estimation error as a function of $\lambda$ in the same setting of Figure~\ref{fig:as_dep} with  $\sigma_\ep =1, \alpha = 1$, and $\gamma = 0.2$.}
    \label{fig:ridge}
  \end{center}
  \vspace{-2cm}
\end{wrapfigure}
We observe that in the high redundancy case where $\omega = 0.2$, the bias starts increasing around $\gamma = 0.5$ whereas in the low redundancy case with $\omega = 0.8$, the bias is almost zero for $\gamma < 1$, changing the critical $\gamma$ after which the model is overparameterized. This is due to the change of the number of effective samples.  The peak of $R(\hat\beta)$ which corresponds to double descent \citep{belkin2019reconciling,hastie2022surprises} also shifts as we change the amount of redundancy.
In Figure~\ref{fig:as_dep}, we consider the same example but set $\lambda = \lambda_\star$, and $\gamma = 2$, and plot $R(\hat\beta)$ as a function of $\omega$. We see that the error of the optimally-tuned ridge estimator is decreasing as the redundancy is decreased.

\vspace{-0.4cm}
\paragraph{Effect of Regularization.}

 In Figure~\ref{fig:ridge}, we again consider the setting of Example 2 in Section~\ref{sec:examples} and set $\sigma_\ep = 1, \alpha = 1,$ and $\gamma = 0.2$. We use Corollary~\ref{cor:final} to plot the estimation error as a function of $\log(\lambda)$. In this plot, different curves correspond to different values of $\omega \in [0,1]$. The dashed vertical line corresponds to $\log(\lambda) = \log(\lambda_\star)$. We observed that as predicted by Theorem~\ref{thm:optimal}, the minimum risk is attained at $\lambda = \lambda_\star$ and the optimal ridge regularize does not depend on the matrix $A$ and is the same for all values of $\omega$.

\vspace{-0.2cm}

\section{Conclusions}
This paper presents a detailed asymptotic analysis of ridge regularized linear regression with linearly dependent data, focusing on the high-dimensional proportional regime. By employing random matrix theory and establishing Gaussian universality, we derived precise characterizations of the estimation error and its dependence on spatio-temporal covariance structures. Our results provide new insights into optimal regularization, overparameterization, and the double descent phenomenon in settings with dependent data. Theoretical results closely align with simulations, highlighting the validity of our findings. 

\section{Acknowledgments}
B.M. would like to thank Ingvar Ziemann, Thomas T.C.K. Zhang, and Bruce D. Lee for stimulating discussions on the topic of statistical estimation with dependent data, during the early phases of this project. The work of B.M. and H.H. is supported by The Institute for Learning-enabled Optimization at Scale (TILOS), under award number NSF-CCF-2112665, and the NSF CAREER award CIF-1943064. This work was also supported by a gift from AWS AI to Penn Engineering's ASSET Center for Trustworthy AI.

\bibliography{SIMOD/references}

\newpage
\appendix
\section{Proof of Theorem~\ref{thm:universality}}
\label{sec:pf:universality}

Let $G= [g_1, \dots, g_n]^\top \in \R^{n \times d}$ be a matrix with entries drawn i.i.d. from  Gaussian distribution with the same mean and variance as $\cD_z$. In this section, we show that for a large class of distributions $\cD_z$, the matrix $Z$ can be replaced with $G$ without changing the limiting value of $R(\hat \beta)$. To prove this theorem, given \eqref{eq:in_terms_of_m}, it is enough to show that $m(\lambda;\gamma)$ is invariant under the change of $\cD_z$ to the corresponding Gaussian. To show this, we will use a Lindeberg exchange argument \citep{lindeberg1922neue,korada2011applications,abbasi2019universality}. Without loss of generality, we assume that $\cD_z$ has a standard variation equal to one. To prove this universality, for any $i \in [n]$  we define the matrix $M_i$ as
\begin{align*}
    M_i := \sum_{k = 1}^{i - 1} e_k g_k^\top + \sum_{k = i+1}^{n} e_k z_k^\top.
\end{align*}
This matrix will help us interpolate the matrix $Z$ and $G$ by replacing the entries row-by-row.
Also, we define $Z_i := M_i + e_i g_i^\top$. Note that $G = Z_n$ and $Z = Z_0$.  Thus, we can write  
\begin{align}
    \label{eq:interpolation}
    \Big|\mathrm{trace}\left(B^\top G^\top \Omega_A G B + \lambda n I_d\right)^{-1} &- \mathrm{trace}\left(B^\top Z^\top \Omega_A Z B + \lambda n I_d\right)^{-1}\Big| \nonumber \\
    &\hspace{-3cm}= \Big|\sum_{i = 1}^{n}\mathrm{trace}\left(B^\top Z_i^\top \Omega_A Z_i B + \lambda n I_d\right)^{-1} - \mathrm{trace}\left(B^\top Z_{i-1}^\top \Omega_A Z_{i-1} B + \lambda n I_d\right)^{-1} \Big| \nonumber \\
    &\hspace{-3cm}\leq \sum_{i = 1}^{n}\Big|\,\mathrm{trace}\left(B^\top Z_i^\top \Omega_A Z_i B + \lambda n I_d\right)^{-1} - \mathrm{trace}\left(B^\top Z_{i-1}^\top \Omega_A Z_{i-1} B + \lambda n I_d\right)^{-1} \Big|.
\end{align}
To prove the theorem, it is enough to show that each term in the above sum is $o(1/n)$. For this, we use that fact that $Z_i := M_i + e_i g_i^\top$ and write
\begin{align*}
    \mathrm{trace}\left(B^\top Z_i^\top \Omega_A Z_i B + \lambda n I_d\right)^{-1} &= \mathrm{trace}\left(B^\top (M_i^\top  + g_i e_i^\top) \;\Omega_A\; (M_i + e_i g_i^\top) B + \lambda n I_d\right)^{-1}\\
    &\hspace{-3cm} =\mathrm{trace}\left(B^\top M_i^\top \;\Omega_A\; M_i B + \lambda n I_d + \omega_i \, B^\top g_i  g_i^\top B + B^\top v_i g_i^\top B + B^\top g_i v_i^\top B\right)^{-1},
\end{align*}
where $\omega_i := e_i^\top \Omega_A e_i \in \R$, and $v_i :=  M_i^\top \Omega_A e_i \in \R^d$. Defining
\begin{align*}
    S_i := \left(B^\top M_i^\top \;\Omega_A\; M_i B + \lambda n I_d\right)^{-1}, \quad U_i := [\;B^\top g_i\; |\, B^\top v_i\,] \in \R^{d \times 2}, \quad \text{and} \quad K_i := \begin{bmatrix}
        0 & 1 \\ 1 & - \omega_i
    \end{bmatrix}, 
\end{align*}
we can use the Woodbury matrix identity to write
\begin{align}
    \label{eq:trace1}
    \mathrm{trace}\left(B^\top Z_i^\top \Omega_A Z_i B + \lambda n I_d\right)^{-1} = \mathrm{trace} ( S_i  ) - \mathrm{trace}\left((K_i + U_i^\top S_i\, U_i)^{-1} (U_i^\top S_i^2 \,U_i)\right).
\end{align}
Also, we note that $Z_{i - 1} = M_i + e_i z_i^\top$, which gives
\begin{align*}
    &\mathrm{trace}\left(B^\top Z_{i-1}^\top \Omega_A Z_{i-1} B + \lambda n I_d\right)^{-1}\\ 
    &\hspace{2cm}= \mathrm{trace}\left(B^\top M_i^\top \;\Omega_A\; M_i B + \lambda n I_d + \omega_i \, B^\top z_i z_i^\top B + B^\top v_i z_i^\top B +  B^\top z_i v_i^\top B\right)^{-1}.
\end{align*}
Hence, similarly, we can write
\begin{align}
    \label{eq:trace2}
    \mathrm{trace}\left(B^\top Z_{i-1}^\top \Omega_A Z_{i-1} B + \lambda n I_d\right)^{-1} = \mathrm{trace} ( S_i  ) - \mathrm{trace}\left((K_i + V_i^\top S_i\, V_i)^{-1} (V_i^\top S_i^2 \,V_i)\right),
\end{align}
in which $V_i$ is given by $V_i := [\; B^\top z_i\; |\, B^\top v_i\,] \in \R^{d \times 2}.$  Consequently, using \eqref{eq:interpolation}, \eqref{eq:trace1}, and \eqref{eq:trace2}, we have
\begin{align}
    \label{eq:diff_tr}
    \Big|\mathrm{trace}\left(B^\top G^\top \Omega_A G B + \lambda n I_d\right)^{-1} &- \mathrm{trace}\left(B^\top Z^\top \Omega_A Z B + \lambda n I_d\right)^{-1}\Big|\nonumber \\
    &\hspace{-3cm}\leq \sum_{i = 1}^{n}\Big|\,\mathrm{trace}\left((K_i + U_i^\top S_i\, U_i)^{-1} (U_i^\top S_i^2 \,U_i) - (K_i + V_i^\top S_i\, V_i)^{-1} (V_i^\top S_i^2 \,V_i)\right) \Big|.
\end{align}
We will next analyze each term in this sum. Defining
\begin{align*}
    &\Delta_1 := ( g_i^\top B S_iB^\top  g_i)(-\omega_i + v_i^\top B S_iB^\top v_i) - (1 + v_i^\top BS_iB^\top  g_i)^2\\[0.2cm]
    &\Delta_2 := (z_i^\top BS_iB^\top\, z_i)(-\omega_i + v_i^\top BS_iB^\top v_i) - (1 + v_i^\top BS_iB^\top z_i)^2,
\end{align*}
the matrix $(K_i + U_i^\top S_i\, U_i)^{-1}$ can be written as
\begin{align*}
    (K_i + U_i^\top S_i\, U_i)^{-1} = \frac{1}{\Delta_1} \begin{bmatrix}
        {- \omega_i + v_i^\top BS_iB^\top \, v_i} & -1 -  g_i^\top BS_iB^\top v_i\\[0.2cm]
        -1 -  v_i^\top BS_iB^\top g_i &  g_i^\top BS_iB^\top g_i
    \end{bmatrix}.
\end{align*}
Similarly, for the matrix $(K_i + V_i^\top S_i\, V_i)^{-1}$ we have
\begin{align*}
    (K_i + V_i^\top S_i\, V_i)^{-1} = \frac{1}{\Delta_2} \begin{bmatrix}
        {- \omega_i + v_i^\top BS_iB^\top v_i} & -1 - z_i^\top BS_iB^\top v_i\\[0.2cm]
        -1 -  v_i^\top BS_iB^\top z_i &  z_i^\top BS_iB^\top  z_i
    \end{bmatrix}.
\end{align*}
We also have
\begin{align*}
    U_i^\top S_i^2 \,U_i = \begin{bmatrix}
        g_i^\top BS_i^2 B^\top g_i  &  g_i^\top BS_i^2 B^\top v_i \\[0.3cm]
         g_i^\top BS_i^2 B^\top v_i  & v_i^\top BS_i^2 B^\top v_i
    \end{bmatrix}, \quad \text{and}\quad V_i^\top S_i^2 \,V_i = \begin{bmatrix}
        z_i^\top BS_i^2 B^\top z_i  & z_i^\top BS_i^2 B^\top v_i \\[0.3cm]
        z_i^\top BS_i^2 B^\top v_i  & v_i^\top BS_i^2 B^\top v_i
    \end{bmatrix}.
\end{align*}
Hence, the $i$-th term in the sum in \eqref{eq:diff_tr} is equal to
\begin{align}
\label{eq:simple_form}
    &\Big|\,\mathrm{trace}\left((K_i + U_i^\top S_i\, U_i)^{-1} (U_i^\top S_i^2 \,U_i) - (K_i + V_i^\top S_i\, V_i)^{-1} (V_i^\top S_i^2 \,V_i)\right) \Big|\nonumber\\[0.2cm]
    &\hspace{0.5cm}= \Bigg|\left(-\omega_i + v_i^\top B S_iB^\top v_i\right) \left[\frac{g_i^\top BS_i^2 B^\top g_i}{\Delta_1} - \frac{z_i^\top BS_i^2 B^\top z_i}{\Delta_2}\right]\nonumber\\[0.2cm]
    &\hspace{1cm}+ (v_i^\top BS_i^2 B^\top v_i) \left(\frac{ g_i^\top BS_i^2 B^\top g_i}{\Delta_1}-\frac{z_i^\top BS_i^2 B^\top z_i}{\Delta_2}\right) \nonumber\\[0.2cm] &\hspace{1.5cm}+  2\left[\frac{( g_i^\top BS_i^2 B^\top v_i)(-1 - v_i^\top BS_i B^\top g_i )}{\Delta_1} - \frac{(z_i^\top BS_i^2 B^\top v_i)(-1 - v_i^\top BS_i B^\top z_i )}{\Delta_2}\right]\;\Bigg|.
\end{align}
Based on the assumption that $\cD_z$ is sub-Gaussian, and using the Hanson-Wright inequality \citep{rudelson2013hanson} and noting that by construction $g_i, z_i \independent S_i$, we have
\begin{align*}
    &z_i^\top BS_i B^\top z_i = \E\,(z_i^\top BS_i B^\top \, z_i) + O(1/\sqrt{n}) = \tr(BS_i B^\top) + O(1/\sqrt{n}),\\[0.2cm]
    & g_i^\top BS_i B^\top g_i = \E\,( g_i^\top BS_i B^\top g_i) + O(1/\sqrt{n}) = \tr(BS_i B^\top) + O(1/\sqrt{n}),\\[0.2cm]
    &v_i^\top BS_i B^\top g_i = \E\,(v_i^\top BS_i B^\top g_i) + O(1/\sqrt{n}) = O(1/\sqrt{n}),\\[0.2cm]
    &v_i^\top BS_i B^\top z_i = \E\,(v_i^\top BS_i B^\top z_i) + O(1/\sqrt{n}) = O(1/\sqrt{n}).
\end{align*}
Also, by a simple order-wise argument, $v_i^\top BS_i B^\top v_i = O(1)$.  This shows that if $\omega_i = O(1)$, then $\Delta_1, \Delta_2 = O(1)$, and setting $\delta := \Delta_2 - \Delta_1$, we have $\delta = O(1/\sqrt{n})$. Similar to the argument above, we have
\begin{align*}
    &n (z_i^\top BS_i^2 B^\top z_i) = n\,\E\,(z_i^\top BS_i^2 B^\top z_i) + O(1/\sqrt{n}) = n\, \tr(BS_i^2 B^\top) + O(1/\sqrt{n}),\\[0.2cm]
    &n(g_i^\top BS_i^2 B^\top g_i) = n\,\E\,(g_i^\top BS_i^2 B^\top g_i) + O(1/\sqrt{n}) = n\,\tr(BS_i^2 B^\top) + O(1/\sqrt{n}),\\[0.2cm]
    &n(v_i^\top BS_i^2 B^\top g_i) = n\,\E\,(v_i^\top BS_i^2 B^\top g_i) + O(1/\sqrt{n}) = O(1/\sqrt{n}),\\[0.2cm]
    &n(v_i^\top BS_i^2 B^\top z_i) = n\,\E\,(v_i^\top BS_i^2 B^\top z_i) + O(1/\sqrt{n}) = O(1/\sqrt{n}).
\end{align*}

\paragraph{Term 1} Using the computations above, $ g_i^\top BS_i^2 B^\top g_i - z_i^\top BS_i^2 B^\top z_i = O(n^{-3/2})$.  With this, we can write
\begin{align*}
    \frac{g_i^\top BS_i^2 B^\top g_i}{\Delta_1} - \frac{z_i^\top BS_i^2 B^\top z_i}{\Delta_2} &= \frac{g_i^\top BS_i^2 B^\top g_i}{\Delta_1} - \frac{z_i^\top BS_i^2 B^\top z_i}{\Delta_1 + \delta}\\[0.2cm]
    &= \frac{g_i^\top BS_i^2 B^\top g_i}{\Delta_1} - \frac{z_i^\top BS_i^2 B^\top z_i}{\Delta_1} \left(1+ O(\delta)\right)\\[0.2cm]
    &=O(n^{-3/2})  +O\left(\frac{(z_i^\top BS_i^2 B^\top z_i)\delta}{\Delta_1}\right) = O(n^{-3/2}).
\end{align*}
Hence, assuming $\omega_i = O(1)$, we have
\begin{align*}
    \left(-\omega_i + v_i^\top BS_i^2 B^\top v_i\right) \left[\frac{g_i^\top BS_i^2 B^\top \tilde g_i}{\Delta_1} - \frac{z_i^\top BS_i^2 B^\top z_i}{\Delta_2}\right] = O(n^{-3/2}).
\end{align*}

\paragraph{Term 2} Using the computations above, $ g_i^\top BS_iB^\top g_i - z_i^\top BS_iB^\top z_i = O(n^{-1/2})$. Recalling that $\delta = O(n^{-1/2})$, and $v_i^\top BS_i^2 B^\top v_i = O(n^{-1})$, we have
\begin{align*}
    (v_i^\top BS_i^2 B^\top v_i) \left(\frac{ g_i^\top BS_iB^\top g_i}{\Delta_1}-\frac{z_i^\top BS_iB^\top z_i}{\Delta_2}\right)  = O(n^{-3/2}).
\end{align*}
\paragraph{Term 3} Similarly, using the fact that $g_i^\top BS_i^2 B^\top v_i$ and $z_i^\top BS_i^2 B^\top v_i$ are both $O(n^{-3/2})$ (using a simple order-wise argument), we get
\begin{align*}
     2\left[\frac{( g_i^\top BS_i^2 B^\top v_i)(-1 - v_i^\top BS_iB^\top g_i )}{\Delta_1} - \frac{(z_i^\top BS_i^2 B^\top v_i)(-1 - v_i^\top BS_iB^\top z_i )}{\Delta_2}\right] = O(n^{-3/2}).
\end{align*}

\paragraph{Putting Everything Together}
Using \eqref{eq:simple_form} we get
\begin{align*}
    &\Big|\,\mathrm{trace}\left((K_i + U_i^\top S_i\, U_i)^{-1} (U_i^\top S_i^2 \,U_i) - (K_i + V_i^\top S_i\, V_i)^{-1} (V_i^\top S_i^2 \,V_i)\right) \Big| = O(n^{-3/2}),
\end{align*}
that alongside \eqref{eq:interpolation}, proves that
\begin{align*}
    \Big|\mathrm{trace}\left(B^\top G^\top \Omega_A G B + \lambda n I_d\right)^{-1} &- \mathrm{trace}\left(B^\top Z^\top \Omega_A Z B + \lambda n I_d\right)^{-1}\Big| \to 0,
\end{align*}
which completes the proof.

\section{Proof of Theorem~\ref{thm:error_with_fixed_point}}
\label{sec:pf:error_with_fixed_point}
The function $\bar m (\lambda;\gamma)$ can be written as
\begin{align*}
    \bar m (\lambda;\gamma)(\lambda) &= \lim_{d/ n \to \gamma} \frac{1}{\lambda n}\mathrm{tr}\left( \left(\frac{X^\top X}{\lambda n} +  I_d\right)^{-1} \right) = \lim_{d/ n \to \gamma} \frac{1}{\lambda n} \mathrm{tr}\left( \left(\frac{1}{\lambda n}B^\top Z^\top \Omega_A\, Z\, B +  I_d\right)^{-1} \right),
\end{align*}
where we defined $\Omega_A := A^\top A$. We note that $m(\lambda)$ is the  trace of a rational function of Gaussian random matrices. To compute it, we follow the recent line of work in high dimensional statistics (see e.g., \cite{disagreement,adlam2019random,adlam2020neural,adlam2020understanding}, etc.) and use the linear pencil method. See \cite[Part 1]{bodin2024random} and \cite[Section A.1]{disagreement} for an overview of the linear pencil technique in random matrix theory and statistics.  In the proof, for a matrix $H \in \R^{k \times k}$, we define the normalized trace as
\begin{align*}
    \bar\tr (H) = \frac{1}{k} \tr(H).
\end{align*}
We also define $(\mathrm{id}\otimes\E \bar\tr)$ as the normalized block-trace; i.e., replacing each block by its trace divided by the size of the block. In this proof, we will not write $\lim_{d/n \to \gamma}$ unless when it is not clear from context.

\subsection{Constructing a Linear Pencil}
We first need to build a linear pencil for this problem; i.e., a block-matrix whose blocks are matrices appearing in the trace and one of the blocks of its inverse is the desired rational function. To do this, we introduce the auxillary variables 
\begin{align*}
    &H_0 :=  \left({B^\top Z^\top {\Omega_A}\, Z\, B}/{\lambda n} +  I_d\right)^{-1}, \quad H_1 := B H_0,\\ 
    &H_2 := ZH_1/\sqrt{\lambda n}, \quad H_3 := {\Omega_A} H_2, \quad \text{and} \quad H_4 := Z^\top H_3/\sqrt{\lambda n},
\end{align*}
to linearize the problem.  Note that the trace $\bar m(\lambda;\gamma)$ can be written as
\begin{align}
    \label{eq:G_0-m}
     \bar m = \frac{1}{\lambda}\; \bar\tr(H_0).
\end{align}
With these, we have $B^\top H_4 +  H_0 = I_d$. Putting all these equations in matrix form, we have
\begin{align}
    \label{eq:linear_pencil}
    \underbrace{\begin{bmatrix}
        I & {0} & {0} & {0} & B^\top\\[0.2cm]
        -B & I & {0} & {0} & {0}\\[0.2cm]
        {0} & -Z/\sqrt{\lambda n} & I & {0} & {0} \\[0.2cm]
        {0} & {0} & -{\Omega_A} & I & {0} \\[0.2cm]
        {0} & {0} & {0} & -Z^\top/\sqrt{\lambda n} & I
    \end{bmatrix}}_{:= {L}}
    \begin{bmatrix}
        H_0\\[0.2cm]
        H_1\\[0.2cm]
        H_2\\[0.2cm]
        H_3\\[0.2cm]
        H_4
    \end{bmatrix} 
    = 
    \begin{bmatrix}
        I\\[0.2cm]
        {0} \\[0.2cm]
        {0} \\[0.2cm]
        {0} \\[0.2cm]
        {0} 
    \end{bmatrix}.
\end{align}
This shows that the matrix $H_0$ is equal to the top left block of ${L}^{-1}$. 
Hence, the matrix ${L}$ is a suitable linear pencil for our problem. The matrix $L$ can be written as
\begin{align*}
    L = I - L_Z - L_{A, B},
\end{align*}
in which
\begin{align*}
    L_Z := \begin{bmatrix}
        {0} & {0} & {0} & {0} & {0}\\[0.2cm]
        {0} & {0} & {0} & {0} & {0}\\[0.2cm]
        {0} & Z/\sqrt{\lambda n} & {0} & {0} & {0} \\[0.2cm]
        {0} & {0} & {0} & {0} & {0} \\[0.2cm]
        {0} & {0} & {0} & Z^\top/\sqrt{\lambda n} & {0}
    \end{bmatrix}, \quad \text{and} \quad L_{A, B} := \begin{bmatrix}
        0 & {0} & {0} & {0} & -B^\top\\[0.2cm]
        B & 0 & {0} & {0} & {0}\\[0.2cm]
        {0} & {0} &0 & {0} & {0} \\[0.2cm]
        {0} & {0} & {\Omega_A} & 0 & {0} \\[0.2cm]
        {0} & {0} & {0} & {0} & 0
    \end{bmatrix}.
\end{align*}
\subsection{Calculating the Interactions of Matrices}
We first augment the matrix $L$ to form the symmetric matrix $\bar{L}$ as follows
\begin{align*}
    \bar L = 
    \begin{bmatrix}
        0 & L^\top \\
        L & 0
    \end{bmatrix} = \underbrace{\begin{bmatrix}
        0 & I \\
        I & 0
    \end{bmatrix} }_{:= \mathbb{I}} - \underbrace{\begin{bmatrix}
        0 & L_Z^\top \\
        L_Z & 0
    \end{bmatrix}}_{:= \bar L_Z}  - \underbrace{\begin{bmatrix}
        0 & L_{A,B}^\top \\
        L_{A,B} & 0
    \end{bmatrix} }_{:= \bar L_{A,B}}.
\end{align*}
We define the matrix $G$ as a matrix with entries equal to the block normalized-traces of  $L^{-1}$; i.e., 
\begin{align*}
    G := (\mathrm{id}\otimes\E \bar\tr) ( L^{-1}).
\end{align*}
Next, we augment the matrix $G$ as 
\begin{align*}
    \bar G = \begin{bmatrix}
        0 & G \\
        G^\top & 0
    \end{bmatrix} &:= \begin{bmatrix}
        0 & (\mathrm{id}\otimes\E \bar\tr) ( L^{-1}) \\
        (\mathrm{id}\otimes\E \bar\tr) (( L^\top)^{-1}) & 0
    \end{bmatrix}\\[0.2cm]
    &= (\mathrm{id}\otimes\E \bar\tr) \begin{bmatrix}
        0 &L^{-1} \\
        ( L^\top)^{-1}& 0
    \end{bmatrix} = (\mathrm{id}\otimes\E \bar\tr) (\bar L^{-1}) = (\mathrm{id}\otimes\E \bar\tr) \left((\mathbb{I} - \bar L_Z - \bar L_{A, B})^{-1}\right).
\end{align*}
Note that the matrix $\bar G$ can be seen as the operator-valued Cauchy transform \cite[Definition 9.10]{mingo2017free} of $\bar L_Z + \bar L_{A, B}$; i.e.,
\begin{align*}
    \bar G = \mathcal{G}_{\bar L_Z + \bar L_{A, B}} (\mathbb{I}).
\end{align*}
A common tool for compute this is the R-transform introduced by  \cite{voiculescu1986addition,voiculescu2006symmetries} which enables the characterization of the spectrum of a sum of asymptotically freely independent random matrices. Because $ L_{A, B}$ is a deterministic matrix, the matrices $ L_Z$ and $ L_{A, B}$ are asymptotically freely independent \cite[Theorem 4.8]{mingo2017free} which enables us to use  the subordination property \cite[Equation 9.21]{mingo2017free} and write $\bar{G} = \mathcal{G}_{ L_Z +  L_{A, B}} (I)$ as 
\begin{align*}
    \bar G = \mathcal{G}_{\bar L_{A, B}} (\mathbb{I}  - \mathcal{R}_{\bar L_{Z}}(\bar G) ).
\end{align*}
Hence, using the definition of Cauchy transform, we have
\begin{align}
    \label{eq:self_consistent}
    \bar G = (\mathrm{id}\otimes\E \bar\tr) \left((\mathbb{I}  - \mathcal{R}_{\bar L_{Z}}(\bar G)  - \bar L_{A, B})^{-1}\right),
\end{align}
in which $\mathcal{R}_{\bar L_{Z}}$ is the R-transform of the matrix $\bar L_{Z}$ and has the form
\begin{align*}
    \mathcal{R}_{\bar L_{Z}}(\bar G) = \begin{bmatrix}
        0 & R^\top\\
        R & 0
    \end{bmatrix}.
\end{align*}
Since the matrix $\bar L_Z$ consists of i.i.d. Gaussian blocks, we can use \cite{far2006spectra} to compute $R$. Using this, the non-zero blocks of $R$ are as follows
\begin{align*}
    [R\,]_{3,4} = \frac{\gamma}{\lambda} [G\,]_{2,5}, \quad \text{and}\quad [R\,]_{5,2} = \frac{1}{\lambda} [G\,]_{4,3}.
\end{align*}
Plugging the R-transform into \eqref{eq:self_consistent}, we get 
\begin{align*}
    G = (\mathrm{id}\otimes\E \bar\tr) \begin{bmatrix}
        I & 0 & 0 & 0 & B^\top\\[0.2cm]
        -B & I & 0 & 0 & 0\\[0.2cm]
        0 & 0 & I & - \frac{\gamma}{\lambda} [G\,]_{2,5} & 0\\[0.2cm]
        0 & 0 & -\Omega_A & I & 0\\[0.2cm]
        0 & - \frac{1}{\lambda} [G\,]_{4,3} & 0 & 0 & I
    \end{bmatrix}^{-1}.
\end{align*}

Looking at the $(1,1)$, $(4,3)$, and $(2, 5)$ blocks of this matrix gives the following set of self consistent equations:
\begin{align*}
    \begin{cases}
        [G\,]_{1,1} = \lambda\,  \bar\tr \left((\lambda I + [G\,]_{4,3} \,\Omega_B)^{-1}\right)\\[0.2cm]
        [G\,]_{4,3} = \lambda  \,\bar\tr \left(\Omega_A(\lambda I - \gamma\, [G\,]_{2,5}\, \Omega_A)^{-1}\right)\\[0.2cm]
        [G\,]_{2,5} = -\lambda\, \bar\tr\left(\Omega_B (\lambda I + [G\,]_{4,3}\,\Omega_B)^{-1}\right)
    \end{cases}
\end{align*}
To simplify these equations, we  define 
\begin{align*}
    \kappa_1 := \frac{\lambda}{\gamma [G\,]_{2,5}}, \quad \text{and}\quad \kappa_2 := \frac{\lambda}{[G\,]_{4,3}}.
\end{align*}
With these, and the fact that $[G\,]_{1,1} = \lambda \bar{m}(\lambda;\gamma)$, the self-consistent equations can be simplified as
\begin{align}
\begin{cases}
    \lambda \bar{m}(\lambda;\gamma) = \kappa_2 \, m_{B}\left(\kappa_2\right)\\[0.3cm]
     \kappa_1 \kappa_2 + \kappa_1^2\,\kappa_2\, m_{A}\left(-\kappa_1\right) + \lambda = 0\\[0.3cm]
     \gamma \kappa_1 \kappa_2 - \gamma \kappa_1 \kappa_2^2 \, m_{B}\left(\kappa_2\right) + \lambda = 0 
\end{cases}
\end{align}
From the third equation, we have
\begin{align*}
    \kappa_1  = \frac{\lambda}{\gamma \kappa_2^2 m_B(\kappa_2) - \gamma \kappa_2},
\end{align*}
and plugging this into the second equation and denoting $\kappa := \kappa_2$, we find
\begin{align*}
    \lambda m_A \left(\frac{\lambda}{\gamma \kappa - \gamma \kappa^2 m_B(\kappa)}\right) + \gamma \kappa \left(\kappa m_B(\kappa)-1\right) \left(1 - \gamma + \gamma \kappa m_B(\kappa)\right) = 0.
\end{align*}
From first equation, we have $\bar{m}(\lambda; \gamma) = \kappa m_B(\kappa)/\lambda,$ which concludes the proof.

\section{Derivatives}
\label{sec:derivative}
In this section, we compute $\frac{\partial m}{\partial \lambda}$ by differentiating the fixed point \eqref{eq:fixed_points}. For simplicity we first define the function $\tilde m_B:\R \to \R$ as
\begin{align*}
    \tilde m_B(x) := \gamma x (1 - x\, m_B(x)), \quad x \in \R.
\end{align*}
The derivative of this function is given by
\begin{align*}
    \tilde m_B'(x) = \gamma - 2\gamma x\, m_B(x) - \gamma x^2\, m_B'(x), \quad x \in \R
\end{align*}
With this, the fixed-point \eqref{eq:fixed_points} can be written as
\begin{align*}
        \lambda m_A \left(\frac{\lambda}{\tilde m_B(\kappa)}\right) + \frac{\tilde m_B^2(\kappa)}{\kappa}  - \tilde m_B(\kappa) = 0.
\end{align*}
Taking derivative with respect to $\lambda$, we arrive at
\begin{align*}
    \frac{\partial \kappa}{\partial \lambda} = \frac{\kappa^2\; \tilde{m}_B^2(\kappa)\;m_A \left(\frac{\lambda}{\tilde m_B(\kappa)}\right) +\lambda \; \kappa^2\; \tilde{m}_B(\kappa)\;  m_A' \left(\frac{\lambda}{\tilde m_B(\kappa)}\right) }{ \lambda^2 \kappa^2 {\tilde m_B'(\kappa)}\; m_A' \left(\frac{\lambda}{\tilde m_B(\kappa)}\right) -\tilde m_B^2(\kappa) \Big[{\big(2\kappa \tilde m_B(\kappa)-\kappa^2\big)\;\tilde m_B'(\kappa) - \tilde m_B^2(\kappa)}\Big]}.
\end{align*}
Using this and recalling that $\lambda \bar{m}(\lambda;\gamma) = \kappa_2 \, m_{B}\left(\kappa_2\right)$, we find
\begin{align*}
    \frac{\partial \bar{m}}{\partial \lambda} = \frac{1}{\lambda} \left(m_B(\kappa) + \kappa m_B'(\kappa)\right) \frac{\partial \kappa}{\partial \lambda} - \frac{\kappa}{\lambda^2} m_B(\kappa).
\end{align*}
Here, everything can be computed given $\kappa$ and $\bar{m}(\kappa)$.

\end{document}